\documentclass{article} 
\usepackage{iclr2027_conference,times}

\usepackage{amsmath,amsfonts,bm}

\def\eqref#1{equation~\ref{#1}}

\def\1{\bm{1}}

\DeclareMathAlphabet{\mathsfit}{\encodingdefault}{\sfdefault}{m}{sl}
\SetMathAlphabet{\mathsfit}{bold}{\encodingdefault}{\sfdefault}{bx}{n}

\usepackage{hyperref}
\usepackage{url}
\usepackage[utf8]{inputenc} 
\usepackage[T1]{fontenc}    
\usepackage{booktabs}       
\usepackage{amsfonts}       
\usepackage{nicefrac}       
\usepackage{microtype}      
\usepackage{xcolor}         

\usepackage{graphicx}
\usepackage{amsmath}
\usepackage{multirow}
\usepackage{wrapfig}
\usepackage{caption}
\usepackage{siunitx}
\usepackage{subcaption}
\usepackage{pgfplots}
\pgfplotsset{compat=1.18}

\title{What Makes a Latent Space Shared? Direct and Recoverable Compatibility in Motion VAEs}

\author{
Iago A.~Brito
\And
Walcy S.~R.~Rios
\And
Lucas A.~Fernandes
\And
Kaiki C.~de Oliveira
\And
Arlindo R.~Galv\~ao Filho
\\[1ex]
Advanced Knowledge Center in Immersive Technologies (AKCIT)\\
Federal University of Goi\'as (UFG)\\
\texttt{\{iagoalves,walcy.rios,lucas\_arruda,kaiki\_camilo\}@discente.ufg.br}\\
\texttt{arlindogalvao@ufg.br}
}

\iclrfinalcopy 
\begin{document}

\maketitle

\begin{abstract}
A shared latent space is often assumed to make representations from different schemas directly interchangeable, yet sharing parameters, dimensionality, or a prior does not determine how corresponding instances are expressed in latent coordinates. We show that this assumption conflates recoverable correspondence, direct compatibility, and decoder reuse. In a controlled multi-schema motion VAE on HumanML3D, even identical inputs yield only $5.9$--$17.7\%$ raw Recall@1 and $0.221$--$0.484$ paired-token cosine, despite correspondence being almost perfectly recoverable after linear calibration. Cycle consistency strengthens recoverability but still leaves direct retrieval and token agreement near zero. In contrast, only $64$ paired clips raise raw Recall@1 to $88.4$--$90.3\%$ and token cosine to approximately $0.82$, making most post-hoc alignment unnecessary, although crossed decoding remains substantially worse than native reconstruction. We further find that reducing measured marginal discrepancy does not imply schema invariance. These results show that a shared backbone or prior does not ensure directly compatible latent coordinates, while limited paired correspondence can substantially improve direct compatibility in the tested setting.
\end{abstract}

\section{Introduction}

Human motion representation is a foundational problem in motion research, providing the interface between physical movement and computational models. The same movement can be described through different skeletons, joint sets, coordinate systems, rotation parameterizations, temporal resolutions, and preprocessing pipelines \citep{mahmood2019amass}, yet learning algorithms operate on these encodings rather than on the physical movement itself. Existing approaches often address this heterogeneity by imposing a common representation or parameterization \citep{guo2022generating,loper2015smpl}, or by providing structural information such as skeletal topology, kinematic structure, or known cross-skeleton correspondences \citep{aberman2020skeleton,lee2023same}. Different schemas, however, may still describe the same underlying movement, and a representation intended to be shared across them should make clear what kind of sharing has actually been achieved. In particular, correspondence may be recoverable through a schema-specific transformation without being expressed directly in compatible coordinates, while compatible coordinates may still be insufficient for one interface's decoder to consume another interface's latent representation. We therefore ask what forms of cross-schema compatibility are induced by a shared learning objective and which additional signals are required to identify a directly interoperable latent space.

We study this question with a deliberately minimal multi-schema variational autoencoder in which each schema retains its native parameterization and connects through schema-specific adapters to a shared encoder--decoder backbone regularized by a common Gaussian prior. Rather than assuming that this construction yields a shared representation, we evaluate the compatibility it actually induces using controlled views of the same HumanML3D motions \citep{guo2022generating}, where correspondence is exact by construction and schema variation can be isolated. We evaluate cross-schema compatibility at four levels. First, we pool each motion into a single latent vector and perform full-gallery retrieval with cosine similarity, testing whether corresponding clips are directly comparable in the learned coordinates. Second, we test post-hoc alignment by freezing the representation model and fitting Procrustes, ridge regression, and canonical correlation analysis (CCA) on training correspondences, then repeating full-gallery retrieval on held-out motions in the aligned spaces to measure whether correspondence is recoverable even when the learned latent coordinates are not directly compatible. Third, for temporally aligned views, we compare synchronized posterior-mean latent tokens at corresponding time steps before pooling the latent sequence into a single clip-level vector, directly testing time-step-level coordinate agreement. Finally, cross-schema decoding asks whether latents produced by one adapter can be consumed by another adapter's decoder, providing a functional test of interoperability. We additionally examine marginal alignment as a separate property, since matching latent distributions need not imply any of these paired forms of compatibility. Because calibrated performance depends on both the mapping family and the number of correspondence pairs used to fit it, we interpret it as a diagnostic of recoverable correspondence rather than evidence that the learned coordinates are intrinsically shared.

The results show that these notions of compatibility separate sharply. Even in the strongest control, where two adapters receive identical motions within the same shared VAE, raw Recall@1 remains only \(5.9\)--\(17.7\%\) across three seeds and synchronized paired-token cosine similarity reaches \(0.221\)--\(0.484\), compared with \(0.015\)--\(0.019\) for mismatched pairs. Fitting CCA on \(20{,}942\) training correspondences with the representation model frozen raises held-out retrieval above \(99.6\%\), showing that the correspondence is almost perfectly recoverable by a calibrated linear comparison even though the two interfaces do not directly express it in the same coordinates. Cycle consistency produces the same qualitative separation, with CCA retrieval again approaching perfection while synchronized token similarity remains close to the mismatched-pair baseline, indicating that round-trip constraints can establish a highly predictable relation between schemas without selecting a common latent frame.

Explicit paired correspondence changes the geometry in a different way. With only \(64\) paired clips for the rotation-based view, raw Recall@1 rises to \(88.4\)--\(90.3\%\) and paired-token cosine reaches approximately \(0.82\), providing substantially stronger evidence of direct coordinate agreement, although crossed decoding remains markedly worse than native reconstruction. Marginal alignment behaves differently again, since reducing the discrepancy between latent distributions does not by itself establish paired correspondence or schema-invariant coordinates. Taken together, these results show that marginal similarity, recoverable correspondence, direct clip- and time-step-level agreement, and decoder interoperability are distinct properties, and that neither a shared backbone nor a shared prior is sufficient to establish all of them. Our experiments isolate these effects using controlled deterministic or lossy views of the same HumanML3D motions, while independently collected, noisy, or asynchronous motion domains remain outside the scope of this study.

\section{Related Work}

\paragraph{Heterogeneous motion representations.}
AMASS \citep{mahmood2019amass} fixes a common schema by converting
heterogeneous motion-capture archives to the SMPL+H body model
\citep{loper2015smpl,romero2017mano}, while HumanML3D
\citep{guo2022generating} defines a canonical root-relative feature
layout with continuous 6D joint rotations, later adopted by
\citep{tevet2022mdm,zhang2023t2mgpt,guo2024momask}. Representation
choices also arise within learned motion bottlenecks, where controlled
comparisons of discrete tokenizers show that quantization design can
materially affect reconstruction quality, semantic preservation, and
token efficiency even when the encoder--decoder architecture and
training protocol are fixed \citep{oliveira2026benchmarking}. When
schemas remain distinct, prior work instead introduces information that
relates them through skeletal topology, semantic body-part
correspondences, or paired motions
\citep{aberman2020skeleton,lee2023same,abdul2017motion,hu2023pose,
yamane2010animating}; these approaches either impose a common
parameterization or provide explicit structure between representations,
whereas we ask what forms of compatibility emerge when schema-specific
interfaces are trained through a shared latent backbone and prior.

\paragraph{Consistency objectives.}
When paired examples are unavailable, cross-domain structure is often
encouraged through cycle consistency, which became a standard signal for
unpaired translation \citep{zhu2017cyclegan} and has also been combined
with shared latent representations \citep{liu2017unsupervised}. In
motion, Neural Kinematic Networks \citep{villegas2018nkn} combine cycle
consistency with adversarial retargeting, while Skeleton-Aware Networks
\citep{aberman2020skeleton} additionally constrain the latent code of a
retargeted motion, but a successful round trip does not by itself identify
the intended intermediate correspondence. Exact cycle-consistent
solutions can remain ambiguous up to distribution-preserving
automorphisms \citep{moriakov2020kernel}, and cycle models may preserve
source information without recovering the intended mapping
\citep{chu2017cyclegan}, motivating our distinction between constraints
that make cross-schema relations recoverable and those that actually
produce directly compatible representations.

\paragraph{Representation alignment and identifiability.}
Distributional alignment by MMD
\citep{gretton2012kernel,long2015learning} or adversarial objectives
\citep{ganin2015unsupervised} does not specify which individual samples
correspond, and domain invariance alone need not guarantee transfer
\citep{zhao2019learning}. Latent identifiability depends on structural
assumptions \citep{locatello2019challenging,khemakhem2020variational},
such as diagonal posteriors, which can favor axis-aligned solutions
\citep{rolinek2019variational}, or paired observations, which can
identify additional shared structure
\citep{von2021self,daunhawer2023identifiability}. Representation
comparison is likewise relative to the transformation and downstream
operation being allowed, since CCA, CKA, and fitted linear maps impose
different invariances
\citep{raghu2017svcca,morcos2018insights,kornblith2019similarity,
roeder2021linear}, while anchor-based methods explicitly construct
relations between otherwise distinct latent spaces
\citep{moschella2022relative,maiorca2023latent}. Recent work shows that geometry-preserving and retrieval-oriented
alignment can favor different solutions \citep{achara2026multi} and
that functional stitching can be directional
\citep{almudevar2026bridging}. We study these distinctions within a jointly trained multi-schema model,
separating distributional similarity, recoverable correspondence,
coordinate agreement, and decoder compatibility.

\begin{figure}[t]
    \centering
    \includegraphics[width=\linewidth]{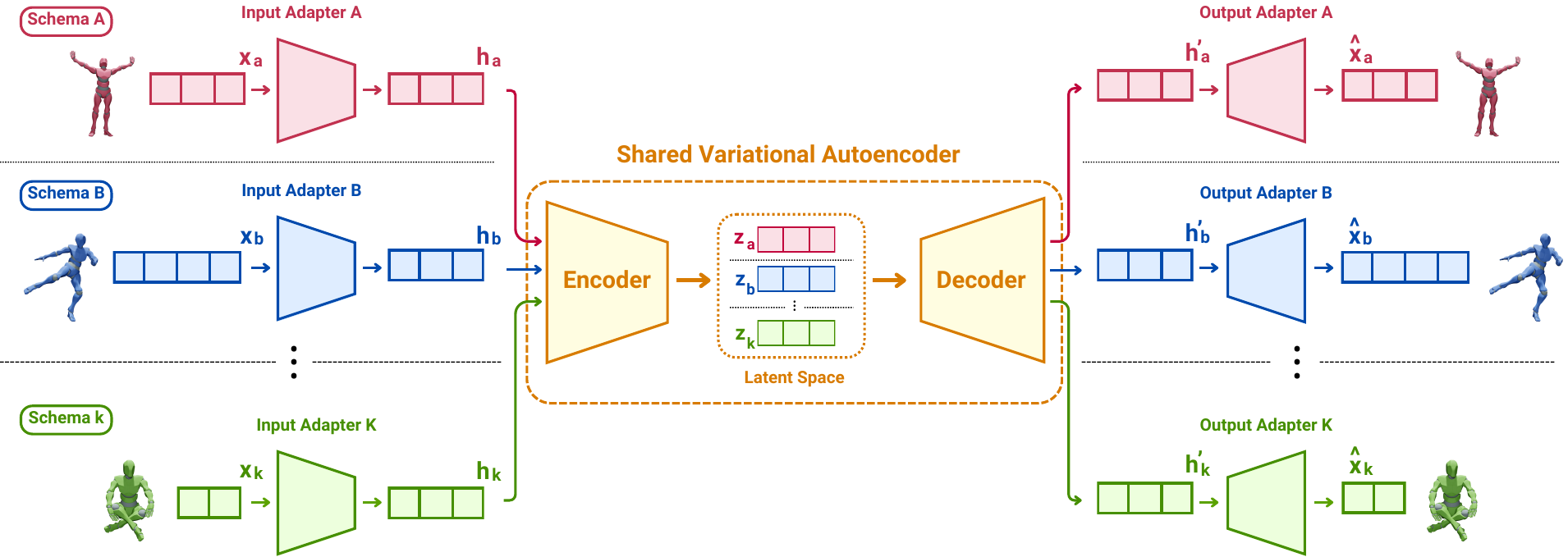}
    \caption{Multi-schema variational autoencoder. Each motion view retains its feature representation and connects to a shared backbone through schema-specific input and output adapters. All schemas share the variational autoencoder and use the same latent dimensionality.}
    \label{fig:pipeline}
\end{figure}

\section{Learning Setting}
\label{sec:setting}

We investigate whether jointly learning motion representations across schemas produces compatible latent representations using a minimal multi-schema variational autoencoder (VAE) \citep{kingma2014autoencoding}. Each schema retains its native representation while schema-specific interfaces connect it to a shared latent backbone, allowing us to vary the cross-schema training signal without changing the basic model construction.

\subsection{Multi-Schema Motion Representation}

Let $\mathcal{D}=\{\mathcal{D}_1,\ldots,\mathcal{D}_N\}$ denote a collection of motion datasets or controlled views. Throughout, $i$ indexes the schema, $n$ the underlying motion instance, and $t$ the time step. In the controlled setting, we begin from underlying motions $\mathcal{Y}=\{y_n\}_{n=1}^{M}$ and construct schema $i$ through a fixed view operator $S_i$,
\begin{equation}
    x_n^{(i)} = S_i(y_n),
    \qquad
    x_n^{(i)}\in\mathbb{R}^{T_n^{(i)}\times d_i},
\end{equation}
where $T_n^{(i)}$ is the number of frames of motion $n$ in schema $i$ and $d_i$ is the schema-specific feature dimension. Thus, $x_n^{(i)}$ and $x_n^{(j)}$ have exact clip-level correspondence because they derive from the same underlying motion $y_n$, even when one view discards information available in the other. This correspondence is known for evaluation but is not necessarily exposed during representation learning, and inputs retain their native feature layout and are standardized independently within each schema.

Schema-specific adapters $A_i^{\mathrm{in}}$ and $A_i^{\mathrm{out}}$ connect each representation to a shared encoder $E$ and decoder $D$, as illustrated in Figure~\ref{fig:pipeline}. The input adapter maps each frame to a common hidden width, while the output adapter returns the decoded features to schema $i$,
\begin{equation}
    x_n^{(i)}
    \xrightarrow{A_i^{\mathrm{in}}}
    h_n^{(i)}
    \xrightarrow{E,\;\mathrm{sampling}}
    z_n^{(i)}
    \xrightarrow{D}
    h_n'^{(i)}
    \xrightarrow{A_i^{\mathrm{out}}}
    \hat{x}_n^{(i)}.
\end{equation}
We write $F_i=E\circ A_i^{\mathrm{in}}$ and $G_i=A_i^{\mathrm{out}}\circ D$ for the effective encoder and decoder.

The architecture is intentionally framewise, with both the adapters and shared encoder--decoder blocks implemented as multilayer perceptrons without temporal mixing, so the temporal length of each sequence is preserved throughout the latent path (complete architectural details are given in Appendix~\ref{app:architecture}). Consequently, token-level comparisons, cycle consistency, and cross-schema decoder reuse are restricted to views defined on the same temporal grid, with the same frame rate and aligned frame times. Views with different frame rates can still be compared through pooled clip-level representations but are excluded from operations requiring one-to-one latent-token correspondence.

For each frame, the effective encoder parameterizes a diagonal-Gaussian posterior,
\begin{equation}
    \left(\mu_{n,t}^{(i)},\sigma_{n,t}^{2(i)}\right)
    =
    F_i\!\left(x_{n,t}^{(i)}\right),
\end{equation}
from which
\begin{equation}
    z_{n,t}^{(i)}
    =
    \mu_{n,t}^{(i)}
    +
    \sigma_{n,t}^{(i)}\odot\epsilon_{n,t},
    \qquad
    \epsilon_{n,t}\sim\mathcal{N}(0,I_{d_z}).
\end{equation}
Stacking the framewise codes gives
\begin{equation}
    z_n^{(i)}
    =
    \left[z_{n,1}^{(i)},\ldots,z_{n,T_n^{(i)}}^{(i)}\right]
    \in\mathbb{R}^{T_n^{(i)}\times d_z},
\end{equation}
with one latent token per input frame. Training objectives use sampled latents, whereas deterministic representation comparisons use the posterior-mean tokens
$m_{n,t}^{(i)}=\mu_{n,t}^{(i)}$.

For clip-level comparisons, we encode the complete motion and summarize its posterior-mean latent sequence by the coordinate-wise temporal mean and population standard deviation,
\begin{equation}
    \bar{m}_n^{(i)}
    =
    \frac{1}{T_n^{(i)}}
    \sum_{t=1}^{T_n^{(i)}}m_{n,t}^{(i)},
    \qquad
    s_n^{(i)}
    =
    \sqrt{
        \frac{1}{T_n^{(i)}}
        \sum_{t=1}^{T_n^{(i)}}
        \left(
            m_{n,t}^{(i)}-\bar{m}_n^{(i)}
        \right)^2
    },
\end{equation}
yielding the fixed-dimensional clip representation
\begin{equation}
    c_n^{(i)}
    =
    [\,\bar{m}_n^{(i)};s_n^{(i)}\,]
    \in\mathbb{R}^{2d_z}.
\end{equation}
Here $s_n^{(i)}$ measures temporal variation of the posterior means rather than posterior uncertainty. The resulting $c_n^{(i)}$ is used for deterministic clip-level retrieval and post-hoc alignment, while synchronized token-level analyses compare the $m_{n,t}^{(i)}$ directly before temporal pooling.

\subsection{Training Objectives}
\label{sec:objective}

All conditions share a common floor of native reconstruction and KL regularization toward a standard Gaussian prior $p$. They differ only in the additional cross-schema information supplied during training, as summarized in Table~\ref{tab:interventions}.

\begin{table}[htbp]
\centering
\caption{Training conditions and their additional cross-schema signals. The discriminator row gives the VAE-side objective, while the domain classifier minimizes $\mathcal{L}_{\mathrm{dom}}$.}
\label{tab:interventions}
\small
\setlength{\tabcolsep}{5pt}
\begin{tabular}{@{}lll@{}}
\toprule
Condition & VAE objective & Additional signal \\
\midrule
Floor
& $\mathcal{L}_{\mathrm{floor}}$
& None \\
Cycle
& $\mathcal{L}_{\mathrm{floor}}
   +\lambda_z\mathcal{L}_{z\mathrm{cyc}}
   +\lambda_x\mathcal{L}_{x\mathrm{cyc}}$
& Round trips \\
MMD
& $\mathcal{L}_{\mathrm{floor}}
   +\lambda_m\mathcal{L}_{\mathrm{MMD}}$
& Latent marginals \\
Discriminator
& $\mathcal{L}_{\mathrm{floor}}
   -\lambda_d\mathcal{L}_{\mathrm{dom}}$
& Schema identity \\
Anchor
& $\mathcal{L}_{\mathrm{floor}}
   +\lambda_a\mathcal{L}_{\mathrm{anchor}}$
& Motion pairs \\
\bottomrule
\end{tabular}
\end{table}

\paragraph{Reconstruction and prior regularization.}
The common floor is
\begin{equation}
    \mathcal{L}_{\mathrm{floor}}
    =
    \mathcal{L}_{\mathrm{rec}}
    +\beta\mathcal{L}_{\mathrm{KL}},
\end{equation}
with
{\small
\begin{equation}
    \mathcal{L}_{\mathrm{rec}}
    =
    \mathbb{E}\!\left[
        \frac{\|x^{(i)}-G_i(z^{(i)})\|_F^2}{T_i d_i}
    \right],
    \qquad
    \mathcal{L}_{\mathrm{KL}}
    =
    \mathbb{E}\!\left[
        \frac{
            D_{\mathrm{KL}}\!\left(
                q_i(z\mid x^{(i)})\,\Vert\,p(z)
            \right)
        }{T_i d_z}
    \right].
\end{equation}
}
The floor jointly trains the adapters and shared parameters without using pair identities, so neither term specifies how the latent coordinate frames of two interfaces should relate.

\paragraph{Cross-schema cycle consistency.}
For $j\neq i$, a sampled source code is decoded through schema $j$ and re-encoded,
\begin{equation}
    \tilde{x}^{(i\rightarrow j)}=G_j(z^{(i)}),
    \qquad
    \tilde{z}^{(i\rightarrow j)}
    \sim q_j(z\mid\tilde{x}^{(i\rightarrow j)}),
\end{equation}
and Cycle penalizes both latent and motion reconstruction error,
{\small
\begin{equation}
    \mathcal{L}_{z\mathrm{cyc}}
    =
    \mathbb{E}\!\left[
        \frac{
            \|\operatorname{sg}(z^{(i)})
            -\tilde{z}^{(i\rightarrow j)}\|_F^2
        }{T_i d_z}
    \right],
    \qquad
    \mathcal{L}_{x\mathrm{cyc}}
    =
    \mathbb{E}\!\left[
        \frac{
            \|x^{(i)}
            -G_i(\tilde{z}^{(i\rightarrow j)})\|_F^2
        }{T_i d_i}
    \right].
\end{equation}
}
The intermediate sequence has no paired reconstruction target, so the objective constrains a cross-interface round trip without identifying which target-schema motion should correspond to the source.

\paragraph{Marginal discrepancy regularization.}
MMD, adversarial regularization, and anchors operate on pooled sampled latents
$r^{(i)}=[\operatorname{mean}_t z_t^{(i)};
\operatorname{std}_t z_t^{(i)}]\in\mathbb{R}^{2d_z}$.
Let $\nu_i$ denote the distribution of $r^{(i)}$ under schema $i$. Squared maximum mean discrepancy \citep{gretton2012kernel} compares these marginals through
\begin{equation}
    \mathcal{L}_{\mathrm{MMD}}
    =
    \mathbb{E}_{i\neq j}
    \left[
        \mathbb{E}k(u,u')
        +\mathbb{E}k(v,v')
        -2\mathbb{E}k(u,v)
    \right],
\end{equation}
where $u,u'\sim\nu_i$ and $v,v'\sim\nu_j$. Thus MMD uses only distributional information and no motion pair identities; the kernel construction is specified in Appendix~\ref{app:objectives}.

\paragraph{Adversarial schema regularization.}
A small MLP domain classifier $H_\phi$ predicts the source schema from pooled latents,
\begin{equation}
    \mathcal{L}_{\mathrm{dom}}
    =
    -\mathbb{E}_{i,x^{(i)},z^{(i)}}
    \left[
        \log H_\phi(i\mid r^{(i)})
    \right].
\end{equation}
The classifier minimizes this loss, while gradient reversal applies $-\lambda_d\mathcal{L}_{\mathrm{dom}}$ to the representation model \citep{ganin2016domain}, discouraging schema-predictive information without supplying motion correspondences.

\paragraph{Paired anchors.}
Anchor is the only primary condition supplied with explicit correspondence between motion instances. For paired windows, symmetric InfoNCE \citep{oord2018representation} encourages the representation of each motion to score above motions with different clip identities,
{\small
\begin{equation}
    \ell_{i\rightarrow j}
    =
    -\frac{1}{B}\sum_{k=1}^{B}
    \log
    \frac{
        \exp\!\left(s_a(r_k^{(i)},r_k^{(j)})/\tau\right)
    }{
        \sum_{\ell\in\mathcal{C}_k}
        \exp\!\left(s_a(r_k^{(i)},r_\ell^{(j)})/\tau\right)
    },
    \qquad
    \mathcal{L}_{\mathrm{anchor}}
    =
    \frac{\ell_{i\rightarrow j}+\ell_{j\rightarrow i}}{2},
\end{equation}
}
where $s_a$ is cosine similarity and $\tau$ is the temperature. Candidate construction, pair sampling, optimization settings, and the paired-attraction ablation are detailed in Appendix~\ref{app:objectives}.

\section{Characterizing Cross-Schema Latent Compatibility}
\label{sec:identification}

\subsection{Direct and Recoverable Correspondence}
\label{sec:shared_properties}

The architecture places latent vectors from every schema in the same ambient space, but this does not imply that corresponding motions are directly compatible. We distinguish correspondence that is already accessible in the learned representations from correspondence recovered only after fitting a transformation on separate motion pairs. At the clip level, \emph{direct retrieval compatibility} means that corresponding motions can be identified under a fixed representation and similarity rule without cross-schema fitting, whereas \emph{recoverable correspondence} refers to comparability obtained only after post-hoc calibration. For synchronized views, we separately use \emph{direct token agreement} to describe numerical similarity between paired posterior-mean latent tokens before temporal pooling.

Marginal similarity provides another distinct notion of sharing, since similar latent distributions do not determine which individual motions correspond. Direct or recoverable correspondence likewise need not imply functional interchangeability across decoders, which we define next. We therefore treat marginal similarity, recoverable correspondence, direct clip- and token-level compatibility, and decoder compatibility as separate operational properties rather than as equivalent consequences of a nominally shared latent space.

\subsection{Decoder Compatibility}
\label{sec:decoder_compatibility}

Direct coordinate agreement does not by itself imply that latent representations are interchangeable across decoders. We call two interfaces \emph{decoder compatible} in the directed $j\rightarrow i$ sense when latent tokens produced by interface $j$ can be passed directly to decoder $G_i$ and recover the corresponding motion in schema $i$. For paired views of motion $n$, we therefore distinguish native decoding,
\begin{equation}
    \hat{x}_{n,\mathrm{nat}}^{(i)}
    =
    G_i\!\left(m_n^{(i)}\right),
\end{equation}
from crossed decoding,
\begin{equation}
    \hat{x}_{n,\mathrm{cross}}^{(j\rightarrow i)}
    =
    G_i\!\left(m_n^{(j)}\right),
\end{equation}
and assess crossed performance relative to the native path against the known schema-$i$ target $x_n^{(i)}$. This is a stronger functional requirement than retrieval or token similarity because the source representation must not only correspond to the target representation, but must lie in a form that the target decoder can use without adaptation. Decoder compatibility is directional and, for lossy views, can also be limited by information absent from the source schema; the Identical setting therefore provides the cleanest control for isolating incompatibility of the latent interfaces themselves.

\subsection{Coordinate Ambiguity and Training Signals}
\label{sec:objective_identification}

The floor objective does not uniquely determine relative latent coordinates across schemas. For a signed permutation $P_i$ applied to every latent frame,
\begin{equation}
    (\mu_i,\Sigma_i,G_i)
    \mapsto
    (P_i\mu_i,\ P_i\Sigma_iP_i^\top,\ G_iP_i^{-1}),
    \qquad
    \Sigma_i=\operatorname{diag}(\sigma_i^2),
\end{equation}
preserves native reconstruction through the compensating decoder transformation. Signed permutations also preserve the diagonal-Gaussian posterior family and unit Gaussian prior, and therefore leave KL regularization unchanged. Thus, at the level of the effective encoder and decoder functions, each schema can admit a different signed permutation without changing the floor objective. Realizing these transformations within the shared architecture requires the closure assumption in Appendix~\ref{app:qinv}; our empirical results do not rely on that assumption.

What matters for cross-schema compatibility is the \emph{relative} coordinate frame. A common signed permutation changes the global frame while preserving pooled cosine comparisons, whereas different permutations across interfaces can leave native reconstruction unchanged while separating corresponding latent tokens and destroying direct comparability. The additional objectives constrain this ambiguity in different ways: cycles impose cross-schema round trips without specifying the intended intermediate motion, MMD and adversarial alignment act on latent marginals, and anchors supply explicit motion correspondences. We therefore test whether these signals produce compatible clip and token representations, either through direct or recoverable correspondence.

\begin{table}[t]
\footnotesize
\centering
\caption{Controlled views of the same HumanML3D motions. Schema A is
native HumanML3D except in Disjoint.}
\label{tab:ladder}
\setlength{\tabcolsep}{8pt}
\begin{tabular}{@{}lllp{0.28\linewidth}@{}}
\toprule
View pair & Schema A & Schema B & Controlled difference \\
\midrule
Identical & native, 263-d & identical, 263-d & none \\
Reordered & native, 263-d & permuted, 263-d & joint/block order \\
Positions & native, 263-d & positions, 71-d & feature selection \\
Rotations & native, 263-d & rotations, 134-d & feature selection \\
Sparse & native, 263-d & sparse, 41-d & joint set and frame subsampling \\
Disjoint & positions, 63-d & rotations, 126-d & disjoint feature blocks \\
\bottomrule
\end{tabular}
\end{table}

\section{Experimental Design}
\label{sec:experimental_design}

We compare the training conditions in Table~\ref{tab:interventions} on paired motion views and evaluate distinct forms of cross-schema compatibility, testing direct clip and token agreement, correspondence recoverable after calibration, latent marginal structure, and cross-schema decoder reuse.

\subsection{Controlled Schema Comparisons}
\label{sec:schema_ladder}

All comparisons use controlled views derived from the same HumanML3D \citep{guo2022generating} motions, providing exact instance correspondence while varying the representation presented to each interface. Schema A is native $263$-dimensional HumanML3D except in \emph{Disjoint}, while schema B varies as shown in Table~\ref{tab:ladder}. Training uses $64$-frame source windows, leaving $20{,}942$ of the $23{,}384$ training clips eligible, whereas retrieval evaluates all $4{,}384$ test clips in full. These controlled views isolate representation differences within one corpus and do not model independently collected, noisy, or asynchronous data sources.

Each view pair defines a separate experiment with one shared VAE and two schema-specific adapter pairs. Pair identities are available for evaluation but are not necessarily exposed during representation learning: Floor, Cycle, MMD, and Discriminator do not use pair identities, whereas Anchor explicitly receives a subset of matched motions. Post-hoc alignment uses correspondences only after the representation model has been frozen. Inputs retain their native feature layout and are standardized independently within each schema. The ladder ranges from identical and reordered inputs to feature-selected, temporally subsampled, and disjoint views; in particular, \emph{Disjoint} compares positional and rotational blocks without shared input channels. Exact feature definitions are given in Appendix~\ref{app:schemas}.

\subsection{Direct and Recoverable Correspondence}
\label{sec:protocol}

For clip-level evaluation, Schema~B queries are ranked by cosine similarity against all $4{,}384$ Schema~A test clips, with Recall@1 as the primary exact-counterpart metric. Raw retrieval uses these representations directly and therefore measures direct clip-level retrieval compatibility; additional ranking and mirror-error analyses are reported in Appendix~\ref{app:ranking}.

For same-rate paired views, we additionally compare synchronized posterior-mean tokens before temporal pooling using paired-token cosine similarity and a symmetric relative distance. Measurements are averaged first within each clip and then equally across clips. No cross-schema transformation or normalization is fitted, and a fixed mismatched-identity control provides a reference for agreement without the true pairing. Metric definitions are provided in Appendix~\ref{app:token_alignment}, with additional token-level results in Appendix~\ref{app:token_results}.

Recoverable correspondence is measured by fitting centered orthogonal Procrustes, ridge regression, or CCA to frozen representations using separate training correspondences, followed by the same full-gallery retrieval. Procrustes permits only an orthogonal transformation of the pooled descriptors, ridge learns a general linear map from schema B to A, and CCA projects both into a common comparison space. CCA uses a fixed maximum of $16$ components, reduced only when required by the fitting sample or representation dimension, and its numerical settings are not selected using test retrieval.

All primary maps and preprocessing statistics are fitted exclusively on training data and applied to the untouched test split. Unless a smaller budget is specified, calibration uses all $20{,}942$ eligible training pairs. Because flexible calibration can also expose identity information in random representations, trained Floor and untrained random encoders are compared under matched $256$-pair and full-pair fitting budgets. Estimator details are given in Appendix~\ref{app:maps}, with complete budget and random-feature controls in Appendix~\ref{app:calibration_controls}.

\subsection{Training Comparisons and Diagnostics}
\label{sec:interventions}

Objective comparisons use \emph{Rotations} unless stated otherwise. Unpaired conditions use unsynchronized windows without pair identities or semantic labels. The main framewise MLP VAE uses $d_z=64$ and $\beta=0.3$, with primary results over three different seeds. The limited-pair Anchor study uses nested budgets of $16$, $64$, $256$, and $1{,}024$ unique training clips and evaluates the final $200{,}000$-step checkpoint to avoid introducing additional paired validation information. At each budget, post-hoc maps fitted to frozen Floor models use the same identities and aligned windows, matching the pair budget and representation granularity but not repeated exposure or training compute. Disjoint is evaluated with Floor, Cycle, and full-pair Anchor, but not with the limited-pair dose study.

Marginal structure is evaluated separately from paired correspondence using a held-out logistic schema probe and a fixed multi-bandwidth RBF MMD kernel whose reference bandwidth is estimated from training representations and held constant across models. Cross-model MMD comparisons use this fixed kernel, while the adaptive training statistic is defined in Appendix~\ref{app:objectives}. Together, the probe and MMD test complementary aspects of marginal structure, and a low kernel discrepancy is not treated as evidence of schema indistinguishability.

Decoder compatibility is also evaluated separately from clip retrieval and token agreement. For selected Identical and Rotations checkpoints, posterior means are decoded either through the native schema A path or from schema B through the schema A decoder on the same $512$ test clips. Local MPJPE is measured over the $21$ non-root RIC joints, excluding root trajectory and orientation. This is a directed $B\rightarrow A$ decoder-reuse test, while training-fitted token maps diagnose whether poor direct reuse is linearly recoverable.

\section{Results}
\label{sec:results}

We report full-gallery $B\rightarrow A$ Recall@1 in percent, with random-ranking chance $0.023\%$. Primary two-interface calibrations use training correspondences only and, unless stated otherwise, are fitted to all $20{,}942$ eligible training clips. Ranges give minima and maxima over three training seeds, rather than confidence intervals.

\subsection{Shared Training Does Not Ensure Direct Clip or Token Compatibility}
\label{sec:shared_results}

The separation between direct and recoverable correspondence appears even when both interfaces receive identical observations. Under Floor, \emph{Identical} reaches only $5.91\%$--$17.72\%$ raw Recall@1, while CCA reaches $99.64\%$--$99.86\%$ (Table~\ref{tab:controlled_results}). This gap is not introduced only by temporal pooling. Synchronized posterior-mean tokens from the same frames have cosine similarity of $0.221$--$0.484$, compared with $0.015$--$0.019$ under mismatched identities, and relative distance $0.989$--$1.227$ versus $1.388$--$1.392$ for the mismatch control. Thus the two interfaces preserve strongly recoverable instance information while assigning corresponding observations to substantially different latent coordinates.

\begin{table}[t]
\centering
\caption{Floor retrieval across three training seeds per view pair. Each experiment has two interfaces sharing one backbone. Maps are fitted separately to all $20{,}942$ eligible training pairs. Entries are Recall@1 percentages; Disjoint uses positional schema A.}
\label{tab:controlled_results}
\small
\setlength{\tabcolsep}{4pt}
\begin{tabular}{@{}lcccc@{}}
\toprule
View pair & Raw & + Procrustes & + Ridge & + CCA-16 \\
\midrule
Identical & 5.91--17.72 & 96.30--98.36 & 97.88--99.11 & 99.64--99.86 \\
Reordered & 0.30--36.82 & 94.37--98.61 & 97.54--99.04 & 99.77--99.91 \\
Positions & 0.11--0.94 & 58.65--61.95 & 60.74--63.59 & 95.78--96.10 \\
Rotations & 0.16--1.64 & 78.85--82.66 & 82.89--87.66 & 95.78--96.30 \\
Sparse & 0.21--0.36 & 44.07--47.35 & 45.28--47.38 & 90.90--91.95 \\
Disjoint & 0.14--0.30 & 19.32--21.26 & 28.60--31.32 & 55.66--57.62 \\
\bottomrule
\end{tabular}
\end{table}

The same pattern holds when the schemas differ. Rotations Floor gives only $0.16\%$--$1.64\%$ raw retrieval and paired-token cosine $0.114$--$0.233$, despite $95.78\%$--$96.30\%$ retrieval after CCA. Reordered also exhibits substantial seed variation in direct compatibility; the seed reaching $36.82\%$ raw retrieval is accompanied by substantially higher token agreement, around $0.71$. Across the evaluated checkpoints, raw Recall@1 and paired-token cosine track one another closely ($\rho_{\mathrm{Spearman}}\approx0.98$), providing descriptive evidence that poor raw retrieval reflects direct token mismatch rather than being created solely by the pooling rule.

Mapped retrieval nevertheless depends strongly on the calibration protocol. With $256$ Rotations training pairs, CCA recovers approximately $67\%$ Recall@1 from trained Floor representations but only about $7\%$ from untrained random encoders. With all $20{,}942$ pairs, the scores rise to approximately $96\%$ and $74\%$, respectively, so the gap narrows as the amount of paired calibration increases. Procrustes remains more selective at the full budget, reaching approximately $82\%$ for Floor versus $37\%$ for random representations. Large-budget CCA therefore demonstrates correspondence accessible to a flexible fitted comparison, but its absolute score is not a standalone measure of learned representation quality. Calibration-budget and random-encoder controls are reported in Appendix~\ref{app:calibration_controls}, with ranking and normalization checks in Appendix~\ref{app:ranking}.

\subsection{Unpaired Constraints Improve Different Properties}
\label{sec:unpaired_results}

Cycle provides the clearest separation between recoverable and direct compatibility. On Rotations it raises Procrustes retrieval to $94.25\%$--$94.89\%$ and CCA to $99.16\%$--$99.25\%$, while raw retrieval remains only $0.09\%$--$0.16\%$ (Table~\ref{tab:unpaired_results}). The same discrepancy is already present before pooling, where paired-token cosine ranges from $-0.025$ to $0.026$ and remains close to the mismatched reference. Cycle therefore regularizes a highly recoverable relation between the interfaces without selecting the same direct coordinate frame.

\begin{table}[t]
\centering
\caption{Unpaired objectives on Rotations across three training seeds. Entries are Recall@1 percentages, with each calibration fitted separately to all eligible training pairs.}
\label{tab:unpaired_results}
\small
\setlength{\tabcolsep}{5pt}
\begin{tabular}{@{}lcccc@{}}
\toprule
Condition & Raw & + Procrustes & + Ridge & + CCA-16 \\
\midrule
Floor
& 0.16--1.64 & 78.85--82.66 & 82.89--87.66 & 95.78--96.30 \\
Cycle
& 0.09--0.16 & 94.25--94.89 & 95.80--96.37 & 99.16--99.25 \\
MMD
& 0.23--2.85 & 80.18--83.07 & 84.24--87.18 & 97.08--97.67 \\
Discriminator
& 12.55--44.14 & 82.66--85.26 & 85.77--88.59 & 96.58--97.42 \\
\bottomrule
\end{tabular}
\end{table}

Marginal objectives affect a different property. Using one fixed evaluation kernel across models, MMD with $\lambda_m=1$ reduces $\mathrm{MMD}^2$ from $0.0130$--$0.0158$ under Floor to $0.0061$--$0.0098$, while raw retrieval remains $0.23\%$--$2.85\%$. The held-out schema probe nevertheless retains $99.95\%$--$99.99\%$ balanced accuracy, with AUC approximately one. The experiment therefore establishes that reducing a measured marginal discrepancy does not imply either schema invariance or paired compatibility; it does not claim that the learned marginals have become indistinguishable.

The discriminator provides the complementary case. Raw retrieval rises to $12.55\%$--$44.14\%$, yet the independent schema probe remains at $99.94\%$--$100\%$ balanced accuracy. Direct counterpart retrieval can therefore improve substantially while schema identity remains readily recoverable (Appendix~\ref{app:marginal_alignment}). Cycle shows the same direct-versus-recoverable separation on Disjoint, where no input channels are shared (Appendix~\ref{app:additional_views}). Additional objective and backbone robustness results are reported in Appendix~\ref{app:ablations}.

\subsection{Paired Correspondence Improves Direct Compatibility but Not Native-Quality Decoder Reuse}
\label{sec:anchor_results}

\begin{figure}[t]
\centering
\includegraphics[width=0.82\linewidth]
{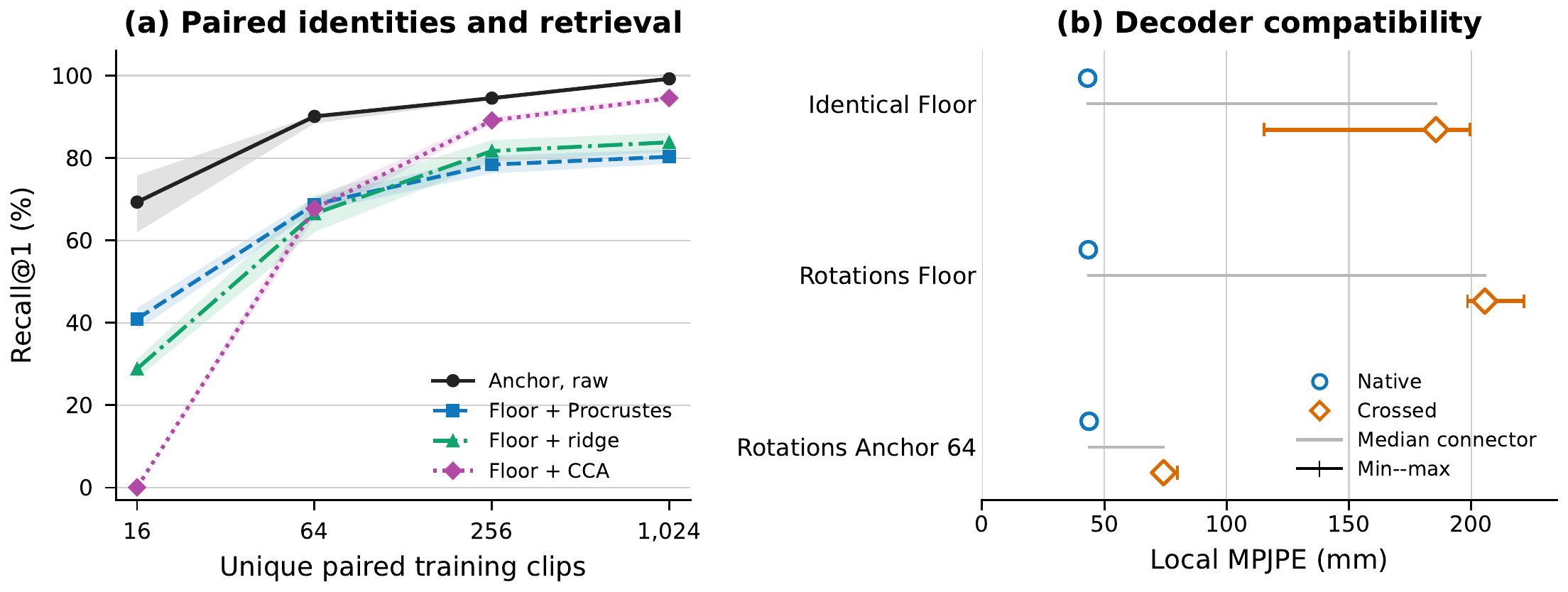}
\caption{Paired supervision and decoder reuse. (a)~Rotations Recall@1 versus paired clip identities. (b)~Native and crossed local MPJPE on the same $512$ clips at matched checkpoints. Lines and markers summarize three runs as described in the text.}
\label{fig:anchor_compatibility}
\end{figure}

Explicit motion correspondences have a markedly different effect. Figure~\ref{fig:anchor_compatibility}(a) shows the Rotations Anchor dose response, with raw retrieval reaching $62.00\%$--$75.75\%$ from $16$ paired training clips, $88.41\%$--$90.33\%$ from $64$, and $94.07\%$--$94.98\%$ from $256$. Post-hoc maps use the same identity budgets and aligned-window granularity, although repeated pair exposure and optimization compute are not matched, so this is not an equal-compute efficiency comparison.

The improvement also appears before temporal pooling. Across the three Anchor-64 runs, paired-token cosine reaches $0.824$--$0.826$, compared with $0.114$--$0.233$ for Rotations Floor, showing that explicit correspondence improves the synchronized latent coordinates rather than only the pooled retrieval descriptor. A paired-MSE ablation yields the same qualitative direct-alignment effect (Appendix~\ref{app:anchor_robustness}).

The limited-pair result is established only for Rotations. With full-pair supervision, Disjoint Anchor achieves $99.59\%$--$99.73\%$ raw retrieval despite having no shared input channels (Appendix~\ref{app:additional_views}). On Rotations, full-pair Anchor achieves perfect raw retrieval and very low measured MMD, while schema identity remains almost perfectly predictable. These results further distinguish direct compatibility from schema invariance.

\paragraph{Decoder compatibility.}
Strong direct compatibility does not ensure native-quality decoder reuse in the tested $B\rightarrow A$ setting. Despite the strong Anchor-64 alignment, crossed MPJPE remains $73.87$--$79.83$~mm versus $43.38$--$44.45$~mm natively (Figure~\ref{fig:anchor_compatibility}(b)). A fitted linear token map substantially improves Floor and Cycle but provides little additional gain for Anchor-64, indicating that its remaining decoder gap is not captured by a simple global linear correction. Complete direct-decoding and token-calibration results are reported in Appendices~\ref{app:cross_decoding} and~\ref{app:token_calibration}, respectively.

\section{Conclusion}

We studied what it means for motion schemas to share a latent
representation when schema-specific interfaces connect them to the same
VAE backbone and prior. Our results separate several properties that
are often conflated: correspondence can be highly recoverable after
calibration despite weak direct clip and token agreement, while explicit
motion pairs substantially improve direct compatibility without ensuring
native-quality cross-decoder reuse. Marginal discrepancy and schema
invariance likewise need not track either form of correspondence. These
conclusions are limited to controlled HumanML3D views, with low-pair
alignment established for Rotations and calibrated recoverability
dependent on the comparison family and correspondence budget. More
broadly, whether a latent space is ``shared'' should be tied to the
operation it is expected to support rather than inferred from shared
architecture, prior, or marginal statistics alone.

\subsection*{AI use statement}

Generative AI tools were used for language editing and proofreading,
including rephrasing text for clarity, and as coding assistants for
implementation tasks. All AI-assisted text and code were reviewed by
the authors, who take full responsibility for the final content of the
paper.


\subsubsection*{Acknowledgments}
This work was fully funded by the project
``Investigação de Tecnologias Imersivas para o Desenvolvimento de Novas
Experiências nas Artes'' of the Advanced Knowledge Center in Immersive
Technologies (AKCIT), with resources from the PPI IoT of the MCTI, grant
057/2023, signed with EMBRAPII. The authors also thank the Fundação de Amparo
à Pesquisa do Estado de Goiás (FAPEG) for financial support (grant 64448878/2024).

\bibliography{iclr2027_conference}
\bibliographystyle{iclr2027_conference}

\appendix


\section{Data and Implementation}
\label{app:implementation}

\subsection{Splits and Controlled Views}
\label{app:schemas}

All views are derived from HumanML3D. Of the $23{,}384$ training and $1{,}460$ validation clips, $20{,}942$ and $1{,}300$, respectively, contain at least $64$ frames and are eligible for training. Evaluation uses all $4{,}384$ test clips, including $2{,}192$ original--mirror pairs, with mirrored motions kept within their official splits.

The native $263$-dimensional representation contains root features (ROOT, $4$-d), root-relative joint positions (POS, $63$-d), continuous 6D rotations (ROT, $126$-d), local joint velocities (VEL, $66$-d), and foot contacts (CONTACT, $4$-d). Positions and rotations describe $21$ non-root joints, while velocities cover all $22$ joints. Table~\ref{tab:app_views} specifies the controlled views.

\begin{table}[htbp]
\centering
\caption{Controlled schema definitions. Each row defines a separate two-interface experiment.}
\label{tab:app_views}
\footnotesize
\setlength{\tabcolsep}{4pt}
\begin{tabular}{@{}llp{0.43\linewidth}r@{}}
\toprule
Comparison & Schema A & Schema B & B width \\
\midrule
Identical & native & Native features in original order & 263 \\
Reordered & native & Permuted joints and feature blocks & 263 \\
Positions & native & ROOT, POS, CONTACT & 71 \\
Rotations & native & ROOT, ROT, CONTACT & 134 \\
Sparse & native & ROOT, 11 non-root positions, CONTACT; every second frame & 41 \\
Disjoint & POS, 63-d & ROT without root or contact features & 126 \\
\bottomrule
\end{tabular}
\end{table}

Reordered applies a fixed permutation of joint groups and feature blocks, preserving the internal order of ROOT and CONTACT. Sparse retains $11$ non-root joints ($\{0,2,4,\ldots,20\}$) and subsamples every second frame without recomputing root increments or contacts. All views use the corresponding HumanML3D normalization statistics, adjusted for channel selection and permutation.

Rotations shares all its input channels with native schema A, whereas Disjoint compares position and rotation features without channel overlap. The latter therefore tests correspondence without shared input features.

\subsection{Architecture}
\label{app:architecture}
\label{app:training}

The model uses schema-specific input and output adapters connected to a shared encoder and decoder. All modules are framewise MLPs, producing one latent token per input frame. Table~\ref{tab:app_architecture} summarizes the architecture. Residual blocks use two linear layers with an intermediate ReLU and an identity skip connection, without normalization or dropout.

\begin{table}[htbp]
\centering
\caption{Main MLP architecture, with hidden width $256$ and latent dimension $64$.}
\label{tab:app_architecture}
\small
\setlength{\tabcolsep}{4pt}
\begin{tabular}{@{}lp{0.68\linewidth}@{}}
\toprule
Module & Layers \\
\midrule
Input adapter
& Linear $d_i\!\rightarrow\!256$, ReLU,
  Linear $256\!\rightarrow\!256$ \\
Shared encoder
& Two width-256 residual blocks, followed by parallel
  mean and log-variance heads $256\!\rightarrow\!64$ \\
Shared decoder
& Linear $64\!\rightarrow\!256$, followed by two
  width-256 residual blocks \\
Output adapter
& Linear $256\!\rightarrow\!256$, ReLU,
  Linear $256\!\rightarrow\!d_i$ \\
Domain classifier
& Linear $128\!\rightarrow\!128$, LeakyReLU,
  Linear $128\!\rightarrow\!128$, LeakyReLU,
  Linear $128\!\rightarrow\!2$ \\
\bottomrule
\end{tabular}
\end{table}

Training uses independently sampled posterior noise at each frame. Reconstruction and auxiliary losses operate on sampled latents, while representation evaluation uses posterior means. Reconstruction is averaged over frames and input features, and the analytical diagonal-Gaussian KL is averaged over frames and latent dimensions.

For the temporal-backbone robustness experiment, the shared MLP residual blocks are replaced with residual Conv1D blocks (kernel size $5$), while the adapters and posterior heads remain framewise. This introduces temporal context and increases the parameter count from approximately $1.04$M to $3.14$M, so the comparison is not parameter-matched.

\subsection{Objective Implementation}
\label{app:objectives}

The objectives are defined in Section~\ref{sec:objective}. Here we describe the implementation choices relevant to their comparison.

\paragraph{Cycle.}
Cycle training averages both interface directions and is restricted to temporally aligned views. A sampled source latent is decoded through the other interface and re-encoded using a fresh posterior sample. Stop-gradient applies only to the reference latent in $\mathcal{L}_{z\mathrm{cyc}}$, while gradients propagate through the cross-schema decoding path. Both cycle losses are averaged over their elements.

\paragraph{MMD.}
MMD is computed on pooled sampled latents using a mixture of five RBF kernels with bandwidth multipliers $\{0.25,0.5,1,2,4\}$. The reference bandwidth is the median pairwise distance in the combined minibatch, treated as constant during differentiation. Within-schema diagonal terms are excluded. Unlike this adaptive training kernel, cross-model evaluation uses a fixed kernel described in Appendix~\ref{app:marginal_protocol}.

\paragraph{Discriminator.}
A domain classifier predicts schema identity from pooled latents. It minimizes $\mathcal{L}_{\mathrm{dom}}$, while gradient reversal applies the opposing gradient, weighted by $\lambda_d$, to the shared encoder and input adapters. The decoder and output adapters receive no direct domain-classification gradient.

\paragraph{Anchor.}
Paired windows are sampled from the same temporal interval of corresponding clips. Symmetric InfoNCE is applied to their pooled sampled latents, excluding other occurrences of the same clip identity from the negative set. The objective updates both input adapters and the shared encoder. The paired-MSE ablation replaces InfoNCE with squared-error attraction between the same paired representations, preserving the sampling protocol.

\subsection{Pair Budgets and Checkpoint Policy}
\label{app:checkpoint_policy}

Limited-pair experiments use nested subsets of $16$, $64$, $256$, and $1{,}024$ training clip identities, fixed independently of optimization seeds. Each budget counts unique paired identities rather than the number of paired windows repeatedly sampled during training.

To avoid introducing additional paired validation information through checkpoint selection, all limited-pair Anchor results use the final checkpoint at $200{,}000$ steps. All-pairs Anchor and the unpaired conditions instead use validation-selected checkpoints unless otherwise stated.


\section{Evaluation and Calibration}
\label{app:evaluation}

\subsection{Clip Representations and Retrieval}
\label{app:retrieval}

As defined in the main text, full-clip representations concatenate the temporal mean and standard deviation of posterior-mean latent tokens, capturing their average activation and temporal variation. Clips are encoded in full, without sliding-window aggregation.

Each schema B test clip is compared against all $4{,}384$ schema A test clips using cosine similarity. Recall@1 is the primary metric, supplemented by Recall@5, Recall@10, mean reciprocal rank, and median rank. The correct match is the exact counterpart of the same source motion; mirrored counterparts count as errors and are examined separately in Appendix~\ref{app:ranking}.

\subsection{Direct Token Agreement}
\label{app:token_alignment}

For temporally aligned views, synchronized posterior-mean tokens are compared directly, without pooling or fitted transformations. We measure paired-token cosine similarity, relative Euclidean distance,
\begin{equation}
    d_{\mathrm{rel}}(a,b)
    =
    \frac{\|a-b\|_2}
    {\sqrt{
        \tfrac{1}{2}(\|a\|_2^2+\|b\|_2^2)+10^{-8}
    }},
\end{equation}
and token RMSE. Metrics are averaged over frames within each clip and then equally across clips, avoiding duration-dependent weighting.

A fixed mismatched-identity pairing of test clips provides a reference for token agreement without true correspondence. The same control is used across checkpoints and is excluded from training, calibration, and model selection.

\subsection{Training-Only Calibration}
\label{app:maps}

Post-hoc calibration is performed with frozen representations using three linear alignment methods. Centered orthogonal Procrustes maps schema B to A through an orthogonal transformation, allowing reflections but not scaling. Ridge regression learns an unrestricted linear map, while CCA projects both representations into a shared canonical comparison space.

Ridge uses a regularization coefficient of
$\alpha_{\mathrm{clip}}=10^{-3}\operatorname{tr}(B_c^\top B_c)/p_B$,
where $B_c$ contains centered schema B descriptors and $p_B$ is their dimension. CCA standardizes each schema, regularizes covariance matrices by $10^{-6}I$, and retains at most $16$ canonical components. All calibrated retrieval scores use cosine similarity in the corresponding comparison space.

Unless otherwise specified, maps and preprocessing statistics are fitted using all $20{,}942$ eligible training correspondences and applied unchanged to the test set. Reduced-budget experiments use subsets of these training pairs. Test-fold calibration is restricted to explicitly identified ancillary sanity checks.

For comparisons with Anchor, calibration uses the same nested paired identities. Window-based fitting uses eight aligned $64$-frame windows per identity, matching the correspondence budget and representation granularity, but not repeated training exposure or computational cost. Calibration-budget and random-encoder controls are reported in Appendix~\ref{app:calibration_controls}.

\subsection{Marginal Diagnostics}
\label{app:marginal_protocol}

Schema separability is measured using an independent logistic-regression probe, separate from the discriminator used during training. Five-fold grouped evaluation keeps original and mirrored motions together, and reports balanced accuracy and AUC.

Cross-model MMD is evaluated using a fixed multi-bandwidth RBF kernel, with its reference bandwidth estimated from training representations of the Rotations Floor reference model. Unlike the adaptive training kernel, this evaluation kernel is held constant across all checkpoints. MMD is estimated on the same disjoint test subsets using the unbiased U-statistic, ensuring that measured discrepancies are comparable across conditions.

\subsection{Decoder and Token-Map Diagnostics}
\label{app:decoder_protocol}

Decoder compatibility is evaluated in the directed $B\rightarrow A$ setting, comparing native reconstruction through $G_A$ with direct decoding of schema B posterior means through the same decoder. Both are evaluated against schema A targets on the same $512$ test clips.

Local MPJPE is measured over the $21$ non-root joints represented by HumanML3D's root-invariant position (RIC) features, after denormalization with schema A statistics. The metric excludes root trajectory and orientation, without forward kinematics or rigid alignment.

To assess whether decoder incompatibility is linearly recoverable, we additionally fit a ridge map between synchronized schema B and A latent tokens using $200{,}000$ training-frame pairs from $6{,}250$ clips. The mapped tokens are then passed to decoder $G_A$. This diagnostic uses substantially more paired identities than Anchor-64 and is not a matched-budget comparison.

\section{Additional Retrieval Results and Controls}
\label{app:additional_results}

\subsection{Per-Seed Retrieval Results}
\label{app:floor_complete}

Tables~\ref{tab:app_floor_complete} and
\ref{tab:app_objective_complete} provide the individual training runs
underlying the main-paper results. All post-hoc maps are fitted to frozen
representations using the $20{,}942$ eligible training correspondences.

\begin{table}[htbp]
\centering
\caption{Floor Recall@1 (\%) across controlled views and training seeds.
Retrieval is full-gallery $B\rightarrow A$; Disjoint uses positions
as schema A.}
\label{tab:app_floor_complete}
\small
\setlength{\tabcolsep}{4pt}
\begin{tabular}{@{}lrrrrr@{}}
\toprule
View pair & Seed & Raw & Proc. & Ridge & CCA \\
\midrule
Identical & 42 & 10.65 & 98.36 & 99.11 & 99.86 \\
Identical & 1  & 5.91  & 96.30 & 97.88 & 99.64 \\
Identical & 2  & 17.72 & 98.02 & 98.77 & 99.86 \\
\midrule
Reordered & 42 & 0.30  & 94.37 & 97.54 & 99.79 \\
Reordered & 1  & 1.46  & 98.15 & 98.77 & 99.77 \\
Reordered & 2  & 36.82 & 98.61 & 99.04 & 99.91 \\
\midrule
Positions & 42 & 0.94 & 58.65 & 60.74 & 95.85 \\
Positions & 1  & 0.11 & 61.95 & 63.59 & 96.10 \\
Positions & 2  & 0.16 & 60.06 & 61.13 & 95.78 \\
\midrule
Rotations & 42 & 0.32 & 82.66 & 87.66 & 96.30 \\
Rotations & 1  & 1.64 & 78.85 & 82.89 & 96.24 \\
Rotations & 2  & 0.16 & 82.30 & 84.92 & 95.78 \\
\midrule
Sparse & 42 & 0.30 & 44.07 & 45.28 & 90.90 \\
Sparse & 1  & 0.21 & 47.35 & 47.38 & 91.95 \\
Sparse & 2  & 0.36 & 46.17 & 45.76 & 91.40 \\
\midrule
Disjoint & 42 & 0.14 & 21.26 & 31.32 & 57.62 \\
Disjoint & 1  & 0.21 & 19.73 & 28.60 & 57.19 \\
Disjoint & 2  & 0.30 & 19.32 & 29.77 & 55.66 \\
\bottomrule
\end{tabular}
\end{table}

\begin{table}[htbp]
\centering
\caption{Rotations objective comparisons across training seeds.
Entries are Recall@1 (\%). MMD uses $\lambda_m=1$;
full-pair Anchor achieves $100\%$ across all three seeds.}
\label{tab:app_objective_complete}
\small
\setlength{\tabcolsep}{4pt}
\begin{tabular}{@{}lrrrrr@{}}
\toprule
Condition & Seed & Raw & Proc. & Ridge & CCA \\
\midrule
Cycle & 42 & 0.11 & 94.37 & 95.80 & 99.16 \\
Cycle & 1  & 0.16 & 94.89 & 96.37 & 99.20 \\
Cycle & 2  & 0.09 & 94.25 & 95.94 & 99.25 \\
\midrule
MMD & 42 & 0.23 & 83.07 & 87.18 & 97.08 \\
MMD & 1  & 0.87 & 82.07 & 86.09 & 97.22 \\
MMD & 2  & 2.85 & 80.18 & 84.24 & 97.67 \\
\midrule
Discriminator & 42 & 44.14 & 85.26 & 87.89 & 96.81 \\
Discriminator & 1  & 12.55 & 82.66 & 85.77 & 96.58 \\
Discriminator & 2  & 17.56 & 82.92 & 88.59 & 97.42 \\
\midrule
Anchor, all pairs & All & 100.00 & 100.00 & 100.00 & 100.00 \\
\bottomrule
\end{tabular}
\end{table}

\subsection{Disjoint-Schema Comparisons}
\label{app:additional_views}

Disjoint compares positional and rotational representations without
shared input channels, providing a stricter test of cross-schema
correspondence than the overlapping feature views. Table~\ref{tab:app_additional_views}
compares Floor, Cycle, and full-pair Anchor under this setting.

\begin{table}[htbp]
\centering
\caption{Disjoint Recall@1 (\%) across three training seeds.
Ranges are min--max; calibration uses all eligible training pairs.}
\label{tab:app_additional_views}
\small
\setlength{\tabcolsep}{4pt}
\begin{tabular}{@{}lrrrr@{}}
\toprule
Condition & Raw & Proc. & Ridge & CCA \\
\midrule
Floor
& 0.14--0.30 & 19.32--21.26 & 28.60--31.32 & 55.66--57.62 \\
Cycle
& 0.14--0.32 & 50.05--55.02 & 60.47--63.05 & 77.08--78.49 \\
Anchor, all pairs
& 99.59--99.73 & 99.61--99.75 & 99.70--99.82 & 99.16--99.57 \\
\bottomrule
\end{tabular}
\end{table}

Cycle substantially improves calibrated correspondence without
improving direct retrieval, even without shared input channels.
Full-pair Anchor instead achieves near-perfect direct retrieval,
showing that explicit correspondence can align these representations
when sufficient paired supervision is available. This experiment
does not establish the low-pair behavior observed on Rotations.

\subsection{Calibration Budgets and Random Features}
\label{app:calibration_controls}
\label{app:geometry}

Calibrated retrieval depends on both the learned representations and
the correspondence pairs used to fit the mapping. To separate these
effects, we compare trained Floor models with untrained random
encoders of the same architecture under matched calibration budgets.
Both conditions use three independently initialized models.

\begin{table}[htbp]
\centering
\caption{Training-only calibration with limited and full
correspondence budgets. Values are median Recall@1 (\%) with
min--max ranges across three seeds. $\Delta$R@1 denotes the
difference between trained and random medians.}
\label{tab:app_random_budget}
\footnotesize
\setlength{\tabcolsep}{3.2pt}
\begin{tabular}{@{}llrccc@{}}
\toprule
View & Map & Pairs & Trained Floor & Random encoder & $\Delta$R@1 \\
\midrule
Rotations & Proc. & 256
& 78.49 (75.57--78.70) & 24.04 (22.31--26.69) & +54.45 \\
Rotations & Proc. & 20,942
& 82.30 (78.85--82.66) & 36.68 (33.10--38.18) & +45.62 \\
\midrule
Rotations & CCA & 256
& 67.24 (67.04--69.53) & 6.84 (6.64--7.82) & +60.40 \\
Rotations & CCA & 20,942
& 96.24 (95.78--96.30) & 73.63 (72.15--73.79) & +22.60 \\
\midrule
Disjoint & Proc. & 256
& 14.80 (13.96--15.31) & 4.43 (2.01--5.09) & +10.38 \\
Disjoint & Proc. & 20,942
& 19.73 (19.32--21.26) & 9.97 (6.41--10.45) & +9.76 \\
\midrule
Disjoint & CCA & 256
& 11.95 (11.59--12.11) & 1.62 (1.48--1.82) & +10.33 \\
Disjoint & CCA & 20,942
& 57.19 (55.66--57.62) & 31.87 (30.52--33.51) & +25.32 \\
\bottomrule
\end{tabular}
\end{table}

With $256$ Rotations pairs, CCA retrieves approximately $67\%$ of
counterparts from trained Floor representations but only $7\%$ from
random encoders. With all $20{,}942$ pairs, random-encoder retrieval
rises to approximately $74\%$, narrowing the gap substantially.
Procrustes remains more selective to learned structure under the
full calibration budget.

These controls show that flexible calibration can recover substantial
instance information even from random representations. High mapped
retrieval is therefore not, by itself, evidence of representation
quality. The trained models nevertheless provide substantially more
recoverable correspondence, particularly when calibration data are
limited.

\subsection{Ranking and Sanity Checks}
\label{app:ranking}

The direct-versus-calibrated separation is not specific to Recall@1.
For Rotations Floor (seed 42), raw retrieval places the correct
counterpart at median rank $1{,}140$, compared with rank $1$ after
either Procrustes or CCA. Mean reciprocal rank increases from
$0.011$ to $0.889$ and $0.980$, respectively.

Simple normalization does not resolve direct incompatibility.
Centering and standardization leave all Identical Floor seeds below
$20\%$ raw Recall@1. In Reordered, standardization improves one
seed from $36.82\%$ to $57.69\%$, still well below calibrated
retrieval.

Known-feature controls clarify the role of shared input information.
In Rotations, direct matching using known channel correspondences
achieves $100\%$ Recall@1 from root features alone and $99.8\%$
from contacts. These controls use explicitly aligned input channels
and do not demonstrate learned latent compatibility. Duration-only
retrieval reaches approximately $4.4\%$.

Shuffling correspondence identities before calibration reduces
retrieval to near chance ($0.00\%$ for Procrustes and $0.02\%$ for
CCA), confirming that fitted mappings exploit genuine pair
structure. This transductive sanity check is separate from the
primary training-only calibration protocol.

Finally, mirror substitutions explain only a small fraction of raw
retrieval failures. For Rotations Floor (seed 42), mirrors account
for $45$ of $4{,}370$ raw errors, compared with $48$ of $162$
CCA errors. Mirror ambiguity therefore becomes more prominent
after calibration but does not explain the poor direct retrieval.

\section{Token Agreement and Marginal Structure}
\label{app:token_marginal}

\subsection{Synchronized Token Agreement}
\label{app:token_results}

Table~\ref{tab:app_token_alignment} reports direct agreement between
synchronized posterior-mean tokens, without pooling or fitted
cross-schema transformations. Metrics are averaged equally across
clips, with ranges over three training seeds.

\begin{table}[htbp]
\centering
\caption{Direct token agreement and raw clip Recall@1 (\%) across
three seeds. The mismatched-identity reference ranges from
$-0.057$ to $0.019$ in cosine similarity and $1.388$ to
$1.447$ in relative $L_2$.}
\label{tab:app_token_alignment}
\footnotesize
\setlength{\tabcolsep}{6pt}
\begin{tabular}{@{}lccc@{}}
\toprule
Condition & Raw R@1 $\uparrow$ & Cosine $\uparrow$ & Rel.\ $L_2$ $\downarrow$ \\
\midrule
Identical Floor
& 5.91--17.72 & 0.221--0.484 & 0.989--1.227 \\
Reordered Floor
& 0.30--36.82 & 0.084--0.706 & 0.706--1.338 \\
Rotations Floor
& 0.16--1.64 & 0.114--0.233 & 1.224--1.319 \\
Rotations Cycle
& 0.09--0.16 & $-0.025$--0.026 & 1.391--1.427 \\
Rotations Anchor-64
& 88.41--90.33 & 0.824--0.826 & 0.562--0.566 \\
\bottomrule
\end{tabular}
\end{table}

Even with identical inputs, Floor produces limited token agreement.
Cycle leaves corresponding tokens close to the mismatched reference
despite nearly perfect calibrated retrieval, whereas Anchor-64
substantially improves direct agreement. These results confirm that
the separation between direct and recoverable correspondence is not
solely an effect of temporal pooling.

\subsection{Marginal Discrepancy and Schema Separability}
\label{app:marginal_alignment}

Table~\ref{tab:app_marginal_audit} compares direct retrieval, held-out
schema classification, and fixed-kernel MMD on Rotations. The latter
uses the same evaluation kernel across models, as described in
Appendix~\ref{app:marginal_protocol}.

\begin{table}[htbp]
\centering
\caption{Marginal diagnostics on Rotations across three seeds.
Lower probe accuracy indicates weaker schema separability; lower
MMD$^2$ indicates smaller measured distributional discrepancy.
MMD$^2$ is reported in units of $10^{-3}$.}
\label{tab:app_marginal_audit}
\footnotesize
\setlength{\tabcolsep}{7pt}
\begin{tabular}{@{}lccc@{}}
\toprule
Condition & Raw R@1 $\uparrow$ & Probe acc. $\downarrow$ & MMD$^2$ $\downarrow$ \\
\midrule
Floor
& 0.16--1.64 & 99.91--99.97 & 13.04--15.78 \\
Cycle
& 0.09--0.16 & 99.93--99.98 & 50.76--63.88 \\
MMD ($\lambda_m=1$)
& 0.23--2.85 & 99.95--99.99 & 6.14--9.81 \\
Discriminator
& 12.55--44.14 & 99.94--100.00 & 16.40--21.54 \\
Anchor (all pairs)
& 100.00 & 99.93--100.00 & 0.36--0.48 \\
\bottomrule
\end{tabular}
\end{table}

MMD regularization reduces the measured discrepancy without
improving direct retrieval or making schemas indistinguishable.
Conversely, Discriminator improves direct retrieval while schema
identity remains readily recoverable. Full-pair Anchor achieves
near-zero measured MMD and perfect direct retrieval, yet retains
almost perfect schema separability.

All evaluated models remain highly schema-classifiable, with probe
accuracy above $99.9\%$. Thus, reducing a measured marginal
discrepancy does not establish schema invariance or paired
compatibility. The experiments do not demonstrate fully matched
marginals with incorrect correspondence.

\section{Paired-Supervision Controls}
\label{app:anchor_robustness}

\subsection{Checkpoint Sensitivity}
\label{app:anchor_dose}

Limited-pair Anchor results use the fixed final checkpoint to avoid
introducing additional paired identities through validation-based
selection. Table~\ref{tab:app_anchor_checkpoints} compares this
policy with validation-selected checkpoints. The differences show
that checkpoint selection affects the measured dose response,
without consistently favoring either policy.

\begin{table}[htbp]
\centering
\caption{Checkpoint sensitivity for nested Rotations Anchor budgets.
Recall@1 (\%) is reported as min--max across three seeds. Primary
limited-pair results use the final checkpoint at $200{,}000$ steps.}
\label{tab:app_anchor_checkpoints}
\small
\setlength{\tabcolsep}{5pt}
\begin{tabular}{@{}rrrr@{}}
\toprule
Clip pairs & Selected step & Selected R@1 & Final R@1 \\
\midrule
16
& 2,000 & 18.23--25.82 & 62.00--75.75 \\
64
& 4,000 & 85.13--86.20 & 88.41--90.33 \\
256
& 6,000 & 98.91--99.20 & 94.07--94.98 \\
1,024
& 12,000--20,000 & 99.89--100.00 & 99.02--99.32 \\
\bottomrule
\end{tabular}
\end{table}

\subsection{Pair-Budget-Matched Calibration}
\label{app:budget_matched}

Table~\ref{tab:app_budget_matched} provides the numerical comparison
behind Figure~\ref{fig:anchor_compatibility}(a). Post-hoc maps are
fitted to frozen Floor representations using the same paired clip
identities as Anchor. Calibration uses either full-clip descriptors
or eight aligned windows per identity, matching the correspondence
budget but not repeated training exposure or optimization compute.

\begin{table}[htbp]
\centering
\caption{Rotations Recall@1 (\%) under matched paired-identity
budgets. Calibration uses three Floor seeds; Anchor uses three
independent runs at fixed final checkpoints. Ranges are min--max.}
\label{tab:app_budget_matched}
\footnotesize
\setlength{\tabcolsep}{4pt}
\begin{tabular}{@{}rlrrrr@{}}
\toprule
Identities & Fit unit & Proc. & Ridge & CCA & Anchor raw \\
\midrule
16 & Clip
& 10.36--15.56 & 15.40--16.97 & 4.56--5.82 & 62.00--75.75 \\
16 & Window
& 38.32--43.57 & 26.48--31.20 & 0.02--0.07 & 62.00--75.75 \\
\midrule
64 & Clip
& 62.16--65.99 & 29.20--43.34 & 41.51--48.59 & 88.41--90.33 \\
64 & Window
& 66.99--70.46 & 62.02--70.96 & 65.42--70.05 & 88.41--90.33 \\
\midrule
256 & Clip
& 75.48--79.20 & 77.69--83.90 & 62.18--71.88 & 94.07--94.98 \\
256 & Window
& 76.25--80.54 & 79.06--84.35 & 88.30--89.90 & 94.07--94.98 \\
\midrule
1,024 & Clip
& 77.78--81.87 & 81.61--86.79 & 93.64--95.35 & 99.02--99.32 \\
1,024 & Window
& 78.56--82.05 & 81.48--86.15 & 93.84--95.23 & 99.02--99.32 \\
\bottomrule
\end{tabular}
\end{table}

Anchor generally achieves stronger direct retrieval than
post-hoc calibration at the same identity budget, particularly
with $16$ and $64$ pairs. This comparison isolates the number
of unique paired identities, not the amount of optimization
performed with them. The unusually weak $16$-identity window
CCA result should not be interpreted as a general limitation
of CCA, given the small number of distinct motions used for fitting.

\subsection{Pair-Subset and Loss Sensitivity}
\label{app:pair_sensitivity}

The choice of paired identities also affects low-budget performance.
Across three $64$-identity subsets at a fixed training seed, raw
Recall@1 ranges from $85.68\%$ to $90.10\%$; with $256$ identities,
it ranges from $94.32\%$ to $95.26\%$. Thus, the dose response
should not be interpreted as a precise sample-complexity estimate.

The result is also not specific to the InfoNCE objective. In
single-seed experiments, varying the InfoNCE weight over
$\{0.1,0.3,1.0\}$ produces $89.55$--$90.44\%$ raw Recall@1
with $64$ pairs and $94.41$--$95.42\%$ with $256$ pairs.
Replacing InfoNCE with paired MSE at $256$ pairs yields
$97.92\%$. This supports the role of explicit correspondence,
without establishing that paired MSE is generally superior.


\section{Objective and Backbone Robustness}
\label{app:ablations}

\subsection{Cycle Components}

Table~\ref{tab:app_cycle_ablation} isolates the latent and motion
cycle terms on Rotations. Both improve calibrated retrieval relative
to Floor, with their combination giving the strongest recovery.
None produces substantial direct compatibility.

\begin{table}[htbp]
\centering
\caption{Cycle-component ablation on Rotations (seed 42).
Full-gallery Recall@1 (\%); maps use all training correspondences.}
\label{tab:app_cycle_ablation}
\small
\setlength{\tabcolsep}{6pt}
\begin{tabular}{@{}lrrr@{}}
\toprule
Condition & Raw & Proc. & CCA \\
\midrule
Floor             & 0.32 & 82.66 & 96.30 \\
Latent cycle only & 0.14 & 88.62 & 98.08 \\
Motion cycle only & 0.09 & 83.90 & 97.49 \\
Both cycles       & 0.11 & 94.37 & 99.16 \\
\bottomrule
\end{tabular}
\end{table}

This single-seed ablation supports the contribution of both terms
in the tested setting, without establishing a general interaction
between them.

\subsection{Temporal Backbone}
\label{app:backbone}

To test whether the findings depend on the framewise MLP, we
repeat the Rotations experiments with the temporal-convolutional
backbone described in Appendix~\ref{app:architecture}.

\begin{table}[htbp]
\centering
\caption{Backbone robustness on Rotations. Full-gallery Recall@1
(\%), reported as min--max over three seeds. The architectures
are not parameter-matched.}
\label{tab:app_backbone}
\footnotesize
\setlength{\tabcolsep}{5pt}
\begin{tabular}{@{}llrrr@{}}
\toprule
Condition & Backbone & Raw & Proc. & CCA \\
\midrule
Floor & MLP
& 0.16--1.64 & 78.85--82.66 & 95.78--96.30 \\
Floor & TCN
& 0.16--1.98 & 93.77--94.48 & 99.50--99.82 \\
\midrule
Cycle & MLP
& 0.09--0.16 & 94.25--94.89 & 99.16--99.25 \\
Cycle & TCN
& 0.07--0.30 & 97.31--98.04 & 99.82--99.91 \\
\midrule
Anchor, all & MLP
& 100.00 & 100.00 & 100.00 \\
Anchor, all & TCN
& 100.00 & 100.00 & 99.98--100.00 \\
\bottomrule
\end{tabular}
\end{table}

The TCN preserves the separation between weak direct retrieval and
strong calibrated correspondence under Floor and Cycle, while
full-pair Anchor again achieves near-perfect direct retrieval.
Because the TCN also increases model capacity, this experiment
tests robustness across architectures rather than isolating the
effect of temporal context.

\section{Decoder Compatibility}
\label{app:decoder_results}

\subsection{Direct Crossed Decoding}
\label{app:cross_decoding}

Table~\ref{tab:app_decoding_ranges} compares native and crossed
decoding at matched checkpoints. Crossed decoding feeds schema B
posterior means directly to the schema A decoder, without latent
calibration. All conditions use the same $512$ test clips and local
MPJPE metric defined in Appendix~\ref{app:evaluation}.

\begin{table}[htbp]
\centering
\caption{Direct decoder compatibility. Raw Recall@1 (\%) and
local MPJPE (mm), reported as min--max across three seeds.}
\label{tab:app_decoding_ranges}
\small
\setlength{\tabcolsep}{4pt}
\begin{tabular}{@{}lrrr@{}}
\toprule
Condition & Raw R@1 & Native MPJPE & Crossed MPJPE \\
\midrule
Identical Floor
& 5.91--17.72 & 42.99--43.36 & 115.20--199.51 \\
Reordered Floor
& 0.30--36.82 & 42.93--43.56 & 65.47--214.70 \\
Rotations Floor
& 0.16--1.64 & 42.98--43.72 & 198.50--221.52 \\
Rotations Cycle
& 0.09--0.16 & 33.08--33.35 & 433.50--509.66 \\
Rotations Anchor-64
& 88.41--90.33 & 43.38--44.45 & 73.87--79.83 \\
\bottomrule
\end{tabular}
\end{table}

Identical isolates decoder incompatibility without differences in
source information, since both interfaces observe the same motion.
Cycle improves native reconstruction but substantially worsens
direct crossed decoding. Anchor-64 reduces this gap, although
crossed decoding remains worse than the native path.

The gap also persists with additional supervision. For Rotations
Anchor at $256$ and $1{,}024$ paired identities, crossed MPJPE
decreases to $69.05$ and $60.30$ mm, respectively, compared with
native errors of $44.51$ and $43.75$ mm. These higher-budget
results are from a single training seed.

\subsection{Token-Level Calibration}
\label{app:token_calibration}

To test whether decoder incompatibility is linearly recoverable,
we fit a ridge map between corresponding latent tokens and decode
the mapped schema B representations through $G_A$.

\begin{table}[htbp]
\centering
\caption{Token-calibrated decoding on Rotations (seed 42).
Local MPJPE (mm). The ridge map uses $200{,}000$ paired
training frames from $6{,}250$ clips.}
\label{tab:app_token_map}
\small
\setlength{\tabcolsep}{5pt}
\begin{tabular}{@{}lrrr@{}}
\toprule
Condition & Native & Direct cross & Ridge cross \\
\midrule
Floor
& 43.72 & 205.59 & 92.08 \\
Cycle
& 33.16 & 496.71 & 82.28 \\
Anchor-64
& 43.38 & 74.23 & 72.48 \\
\bottomrule
\end{tabular}
\end{table}

Token calibration substantially improves crossed decoding for
Floor and Cycle, indicating that useful decoder-level information
remains recoverable despite poor direct reuse. For Anchor-64,
the improvement is small, suggesting that its remaining decoder
gap is not resolved by the fitted linear transformation.

This is a functional recoverability diagnostic rather than a
matched-budget comparison, since token calibration uses
substantially more paired identities than Anchor-64.

\section{Coordinate Ambiguity and Cycle Consistency}
\label{app:qinv}

\subsection{Floor Invariance and Architectural Scope}

Consider the effective posterior $q_i$ and decoder $G_i$ of
interface $i$. For any signed permutation $P_i$ applied to each
latent token, define
\begin{equation}
\mu_i'=P_i\mu_i,
\qquad
\Sigma_i'=P_i\Sigma_iP_i^\top,
\qquad
G_i'(z')=G_i(P_i^{-1}z').
\label{eq:app_posterior_transform}
\end{equation}

Coupling posterior samples through $z'=P_i z$ preserves native
reconstruction exactly. Since signed permutations preserve the
diagonal-Gaussian posterior family and the standard Gaussian
prior, KL regularization is also unchanged. Thus, independent
interface reparameterizations preserve the Floor objective at
the level of effective posterior--decoder functions.

This invariance does not automatically hold in the parameter
space of a shared neural architecture. Exact realizability of
independent $P_i$ requires the jointly representable effective
functions to remain closed under the transformations in
Equation~\ref{eq:app_posterior_transform}, which shared nonlinear
parameters need not guarantee.

The argument therefore identifies an ambiguity permitted by
the objective under this architectural condition, rather than
proving that all such transformations are realizable or that
learned representations differ only by signed permutations.
A common permutation changes the global coordinate frame,
whereas independent permutations can alter relative
cross-interface agreement.

\subsection{Stochastic Cycles and Correspondence}

Cycle consistency constrains round trips but does not specify
the intended intermediate correspondence. Because the
re-encoded posterior is sampled, even an identity mean
round trip incurs a variance-dependent residual:
\begin{equation}
\mathbb{E}
\left[
\|z-\widetilde z\|_2^2 \mid z
\right]
=
\|z-m_j(z)\|_2^2
+
\operatorname{tr}V_j(z),
\label{eq:app_stochastic_cycle}
\end{equation}
where $m_j(z)$ and $V_j(z)$ are the mean and covariance of
the cross-interface re-encoded posterior. Stop-gradient
changes the derivative, not this expected residual.

To illustrate the remaining ambiguity, consider two
identical schemas with $x\sim\mathcal N(0,I_d)$ and
$d=d_z$. For $0<\eta<1$, let $a=\sqrt{1-\eta^2}$ and
choose arbitrary signed permutations $P_i$. Define
\begin{equation}
q_i(z\mid x)
=
\mathcal N(aP_i x,\eta^2I_d),
\qquad
G_i(z)
=
a^{-1}P_i^\top z.
\label{eq:app_cycle_example}
\end{equation}

Each interface has aggregate posterior
$\mathcal N(0,I_d)$, regardless of $P_i$. For independent
standard-Gaussian noise $\epsilon,\epsilon'$,
\begin{equation}
z=aP_i x+\eta\epsilon,
\qquad
\widetilde z
=
aP_jG_j(z)+\eta\epsilon'
=
z+\eta\epsilon'.
\end{equation}

The expected per-element native reconstruction, latent-cycle,
and motion-cycle losses are respectively
$\eta^2/a^2$, $\eta^2$, and $2\eta^2/a^2$, while the
per-input KL is also independent of the permutations.
Nevertheless, direct cross-schema decoding gives
\begin{equation}
G_j(aP_i x)=P_j^\top P_i x,
\end{equation}
which need not equal the paired target $x$.

Thus, identical latent marginals and consistent stochastic
round trips can coexist with incorrect cross-schema
correspondence. This construction illustrates an ambiguity
of the effective objectives, not a claim about the global
optima or exact realizability of the shared neural architecture.

\subsection{Pooling and Coordinate Comparisons}
\label{app:pooling_symmetry}

The clip representation concatenates temporal means and
coordinate-wise standard deviations. Means commute with
linear transformations, but coordinate-wise standard
deviations generally do not, since transformed variances
depend on cross-coordinate covariance. Consequently,
an orthogonal mapping between pooled descriptors need
not correspond to an orthogonal mapping between latent
tokens.

Signed permutations are an exception. For the pooling
operator $\rho$ and a signed permutation $P$,
\begin{equation}
\rho(Pz)
=
\begin{bmatrix}
P & 0\\
0 & |P|
\end{bmatrix}
\rho(z),
\label{eq:app_pooling_signed_permutation}
\end{equation}
where $|P|$ denotes the entrywise absolute value.
The resulting block transformation is orthogonal,
so a common signed permutation preserves cosine
similarity between pooled representations.

This distinction motivates evaluating direct token
agreement separately from clip-level retrieval.
Calibrated retrieval characterizes correspondence
recoverable under a specified mapping family and
pair budget, but does not establish agreement of the
underlying framewise coordinates or decoder
interchangeability.

\section{Limitations and Future Work}
\label{app:limitations}

Our experiments use controlled views of HumanML3D, providing exact
correspondence but not capturing the noise, temporal misalignment, or
distribution shifts of independently collected motion datasets.
Limited-pair alignment is demonstrated only for Rotations, which
shares input channels with Native, while Disjoint is evaluated with
full-pair supervision. Calibrated retrieval depends on the mapping
family and correspondence budget, and decoder compatibility is
evaluated only in the $B\rightarrow A$ direction. Although the
main findings persist with a temporal backbone, the architectures
are not parameter-matched, and the contribution of backbone sharing
is not isolated.

Future work should examine these distinctions across heterogeneous
skeletons and independently collected motion data, particularly
under limited paired supervision. Further directions include
objectives that explicitly encourage decoder interoperability and
identifying the structural assumptions needed to establish
compatible latent coordinates without paired observations.

\end{document}